\documentclass[runningheads]{llncs}
\usepackage{graphicx}
\begin{document}
\title{One-stage Shape Instantiation from a Single 2D Image to 3D Point Cloud\thanks{Xiao-Yun Zhou and Zhao-Yang Wang contribute equally to this paper. This work was supported by EPSRC project grant EP/L020688/1.}}
\titlerunning{One-stage Shape Instantiation from a Single 2D Image to 3D Point Cloud}
%
\author{Xiao-Yun Zhou\inst{*} \and Zhao-Yang Wang\inst{*} \and Peichao Li \and Jian-Qing Zheng \and Guang-Zhong Yang}
%
\authorrunning{X.-Y. Zhou and Z.-Y. Wang et al.}
%
\institute{The Hamlyn Centre for Robotic Surgery, Imperial College London, UK \email{xiaoyun.zhou14@imperial.ac.uk}}
\maketitle              
\begin{abstract}
Shape instantiation which predicts the 3D shape of a dynamic target from one or more 2D images is important for real-time intra-operative navigation. Previously, a general shape instantiation framework was proposed with manual image segmentation to generate a 2D Statistical Shape Model (SSM) and with Kernel Partial Least Square Regression (KPLSR) to learn the relationship between the 2D and 3D SSM for 3D shape prediction. In this paper, the two-stage shape instantiation is improved to be one-stage. PointOutNet with 19 convolutional layers and three fully-connected layers is used as the network structure and Chamfer distance is used as the loss function to predict the 3D target point cloud from a single 2D image. With the proposed one-stage shape instantiation algorithm, a spontaneous image-to-point cloud training and inference can be achieved. A dataset from 27 Right Ventricle (RV) subjects, indicating 609 experiments, were used to validate the proposed one-stage shape instantiation algorithm. An average point cloud-to-point cloud (PC-to-PC) error of 1.72mm has been achieved, which is comparable to the PLSR-based (1.42mm) and KPLSR-based (1.31mm) two-stage shape instantiation algorithm.
\end{abstract}
\section{Introduction}
Minimally-invasive and robot-assisted surgeries made significant advances in recent years. Compared to traditional open surgery, they decrease patient trauma and shorten the recovery time for patients. However, this development brings challenges to surgeons, as the operation environment is often not visually accessible. Clinicians need to perform very difficult interventions under poor, low-resolution and less-informative 2D navigation. Therefore, shape instantiation which instantiates the 3D shape of a dynamic target from limited 2D images is important for mitigating this challenge. To achieve real-time navigation intra-operatively, there are two main requirements on shape instantiation algorithms: 1) the number of intra-operative 2D images should be kept minimal, otherwise it increases the input collection time and hence is harmful for real-time navigation; 2) fast inference speed during the application.

Shape instantiation is usually based on multiple 2D images, for example, six to seven prostate 2D images were needed in~\cite{cool20063d} for the reconstruction. In~\cite{toth2015adaption}, to predict intra-operative 3D Abdominal Aortic Aneurysm (AAA) deformation caused by catheter insertion, 3D AAA shape was instantiated from two 2D fluoroscopic images with the as-rigid-as-possible method for navigation in Endovascular Aortic Repair (EVAR). Currently, many applications are focusing on using a single 2D intra-operative image as the input. The 3D shape of a fully-deployed, partially-deployed and fully-compressed stent graft was predicted from a single 2D intra-operative fluoroscopic image in~\cite{zhou2018real},~\cite{zheng2019real} and~\cite{zhoustent} respectively for navigating Fenestrated Endovascular Aortic Repair (FEVAR). The 3D AAA skeleton was instantiated from a single intra-operative 2D fluoroscopic image with graph matching and skeleton deformation for FEVAR robotic path planning~\cite{zheng20183d}. The 3D shape of a liver was instantiated from a single 2D image scanned at an optimal scan plane with Statistical Shape Model (SSM) and Partial Least Square Regression (PLSR)~\cite{lee2010dynamic}.
\begin{figure}
\centering
\includegraphics[width=1.0\textwidth]{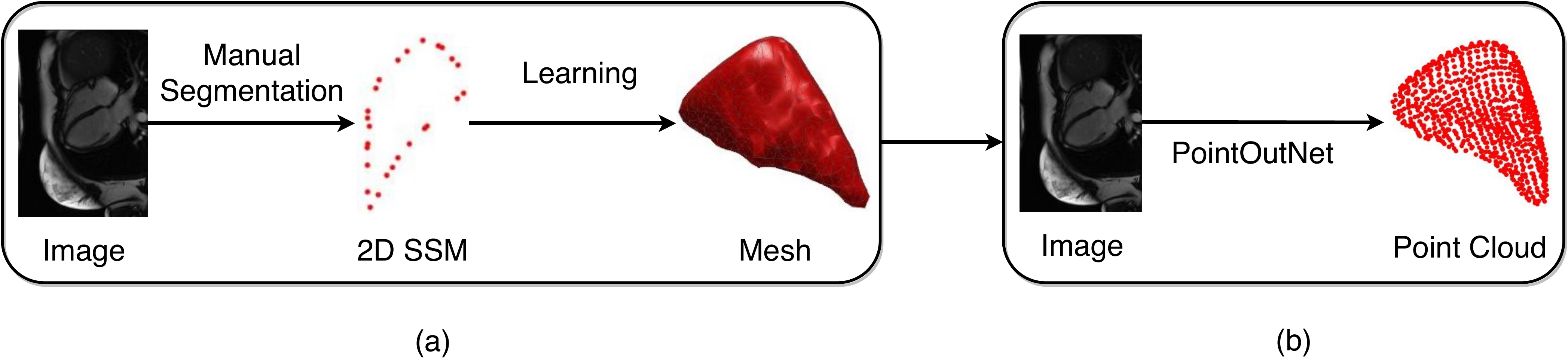}
\caption{Shape instantiation of (a) two-stage with manual image segmentation to generate 2D SSM and KPLSR-based learning for 3D mesh prediction; and (b) one-stage with PointOutNet to predict 3D point cloud from a single 2D image.}
\label{fig:intro}
\end{figure}

A general dynamic framework was proposed recently in~\cite{zhou2018realmed} for 3D shape instantiation. First, it determined an optimal scan plane for a dynamic target by analyzing its pre-operative 3D SSM with Sparse Principal Component Analysis (SPCA). Second, Kernel Partial Least Square Regression (KPLSR) was used to learn the relationship between the pre-operative 2D and 3D SSM. Third, during the inference, with a new intra-operative 2D image at the same optimal scan plane as the input, the KPLSR-learned model was applied to predict the intra-operative 3D shape. However, in~\cite{zhou2018realmed}, the shape instantiation is two-stage with manual segmentation to generate 2D SSM first and then KPLSR-based learning for 3D mesh prediction, as shown in Figure~\ref{fig:intro}(a).

Recently, new neural network architectures were proposed to recover 3D shape from a single 2D image. In \cite{choy20163d}, based on the Long Short-Term Memory (LSTM) framework, a 3D Recurrent Reconstruction Neural Network (3D-R2N2) was proposed to utilize one or more images from different viewpoints to recover a 3D volume. In \cite{fan2017point}, a point cloud prediction network called PointOutNet is proposed to generate un-ordered 3D point cloud from a single RGB image. In \cite{mandikal20183d}, a two-stage architecture called 3D-LMNet was proposed to improve 3D point cloud reconstruction: a 3D point cloud auto-encoder was used to learn the 3D point cloud latent space and then an image encoder was used to map the 2D image to the learned latent space. \cite{kulon2019single} proposed an approach recovering 3D hand mesh from a single image. Their architecture consists of image encoder, graph-convolution-based mesh auto-encoder and decoder.

In this paper, the previously designed shape instantiation is improved to be one-stage, as shown in Figure~\ref{fig:intro}(b). PointOutNet~\cite{fan2017point} with 19 convolutional layers, three fully-connected layers and Chamfer loss is used. The Right Ventricles (RVs) of 27 subjects, including both asymptomatic subjects and subjects with  Hypertrophic Cardiomyopathy (HCM), were used to validate the one-stage shape instantiation with 609 experiments. The results show that the proposed one-stage shape instantiation algorithm can achieve comparable accuracy to previous two-stage ones based on PLSR or KPLSR learning, indicating potential end-to-end inference capability during practical applications.
\section{Methodology}
\subsection{Framework Introduction}
For a dynamic target, i.e. the RV, multiple 3D scans are acquired at different time frames in the dynamic cycle. A 3D mesh is reconstructed for the target at each time frame. For different time frames of a patient, these meshes are expressed into the same number of vertices and the same connectivity $\textbf{Y}_{\rm M\times numY \times 3}$ - 3D SSM with the method in~\cite{nonrigidICP}, where $\rm M$ is the number of time frame, $\rm numY$ is the number of vertices, 3 represents the $(x,y,z)$ coordinate of each vertex. Synchronized 2D images $\textbf{I}_{\rm M\times H \times W}$ at all time frames are acquired at the optimal scan plane, where $\rm H$ is the image height while $\rm W$ is the image width. PointOutNet is trained from $\textbf{I}$ to $\textbf{Y}$. In \cite{zhou2018realmed}, the predicted $\hat{\textbf{Y}}$ by KPLSR is with the same corresponding vertex order as the ground truth $\textbf{Y}$. In this paper where PointOutNet is used, restricting the predicted point cloud with the same vertex order as the ground truth leads to an unsatisfactory result. Hence Chamfer distance is used as the loss function and the predicted $\hat{\textbf{Y}}$ is a point cloud with different vertex order as $\textbf{Y}$. During the inference, with a new intra-operative 2D input image  $\textbf{i}_{\rm H \times W}$, the intra-operative 3D RV point cloud will be predicted by the trained PointOutNet model as $\textbf{y}_{\rm numY \times 3}$.

\subsection{PointOutNet}
\begin{figure}
\centering
\includegraphics[width=1.0\textwidth]{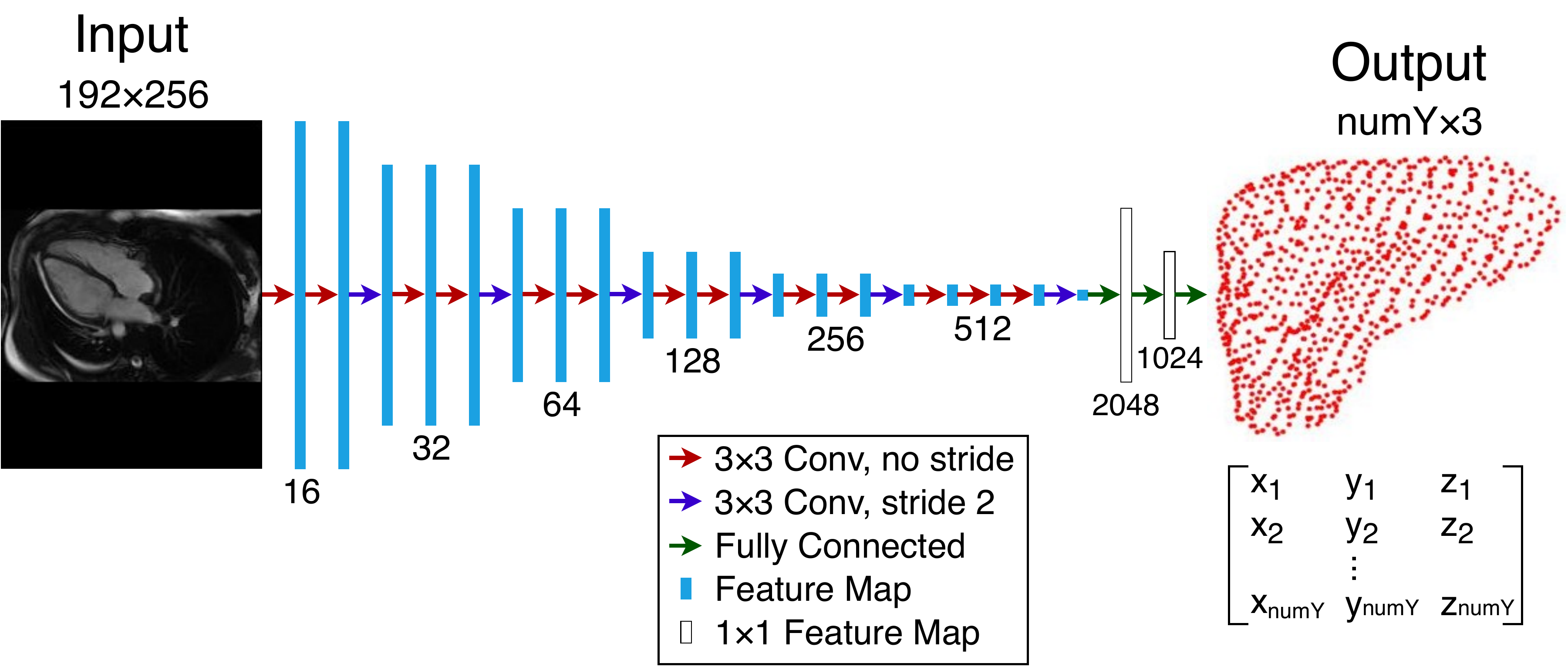}
\caption{Detailed network architecture of PointOutNet.}
\label{fig:network}
\end{figure}

PointOutNet consists of an encoding part with convolutional layers to extract information from images and a prediction part with fully-connected layers to predict vertex coordinates. With an input of $\textbf{F}_{\rm N\times H \times W \times \rm C^{in}}^{\rm in}$, where $\rm N$ is the batch size, $\rm C^{in}$ is the input feature channel, multiple trainable convolutional kernels $\rm \textbf{T}_{\rm C^{in} \times \rm K \times \rm K \times \rm C^{out}}$ move along the input with a stride of $\rm S$, resulting in an output feature map:

\begin{equation}
\rm \textbf{F}_{\rm N\times \rm H'\times \rm W' \times \rm C^{out}}^{out} = \textbf{F}_{\rm N\times \rm H\times \rm W \times \rm C^{in}}^{in} \cdot  \textbf{T}_{\rm C^{in} \times \rm K \times \rm K \times \rm C^{out}}+\textbf{B}_{\rm C^{out}}
\end{equation}
where $\rm K$ is the kernel size, $\rm C^{out}$ is the output feature channel number, $\rm H'=H//S$, $\rm W'=W//S$, $//$ is floor division, $\textbf{B}_{\rm C^{out}}$ is the bias. In the encoding part, ReLU is used as the activation function. Convolutional layers with strides of 2 are used for feature map down-sampling. In order to avoid over-fitting, L2 regularization is applied after each convolutional layer. The channel number of the first convolutional layer is 16 and incrementally doubles after each down-sampling convolution. The output of the encoding part is received by the prediction part which consists of three fully-connected layers with ReLU activation functions and L2 regularizers. The output dimension of each fully-connected layer is 2048, 1024 and $\rm numY\times 3$ respectively. The final output layer uses a linear activation function. Details of the PointOutNet can be found in Figure \ref{fig:network}.

As point cloud is an unordered data format, Using the regular L1 or L2 loss to calculate the corresponding distance error between the predicted point cloud and the ground truth may cause regression difficulty. Hence Chamfer distance is used as the loss function. It calculates the distance between the predicted point cloud and the ground truth as:
\begin{equation}
    Loss = \sum\limits_{\hat{\textbf{y}}\in\hat{\textbf{Y}}} \rm min_{\textbf{y}\in\textbf{Y}}\vert \vert \hat{\textbf{y}} - \textbf{y}\vert \vert_2^2 + \sum\limits_{\textbf{y}\in\textbf{Y}} min_{\hat{\textbf{y}}\in\hat{\textbf{Y}}}\vert \vert \textbf{y} - \hat{\textbf{y}}\vert \vert_2^2
\end{equation}
The advantage of this loss function is that it is easy to differentiate and robust against outliers \cite{fan2017point}.

\subsection{Experimental Setup}
\paragraph{Data collection}
27 RV subjects including 9 Hypertrophic Cardiomyopathy (HCM) patients and 18 asymptomatic subjects were scanned by a 1.5T Magnetic Resonance (MR) scanner (Sonata, Siemens, Erlangen, Germany) from the atrioventricular ring to the apex. The time frames are $19-25$. The slice gap is 10mm. The pixel spacing is $1.5-2mm$. Pre-operative 3D RV meshes were segmented manually from multiple short-axis 2D images for generating pre-operative 3D SSM. Synchronized 2D images were scanned at the four-chamber long axis scan plane which is the optimal scan plane based on~\cite{zhou2018realmed}. Leave-one-out cross validation was performed as suggested by \cite{zhou2018realmed} where one time frame was used for the testing while all others were used for the training. In total, 609 experiments were performed.
\paragraph{Data pre-processing}
The 2D input images were of dimensions of $192\times 256$. Their intensities were normalized to [0, 1] according to the maximum intensity of each image. Batch size was set as one. Batches were fed into the network in a random order. The number of vertices for each patient varied from 800 to 2500, hence the feature channel number in the last layer of the PointOutNet needs to be adjusted individually for each patient. $\textbf{Y}$ was centered at (0, 0, 0) by subtracting its center coordinate. 
\paragraph{Training configuration}
1500 epochs were trained for each experiment. The learning rate was set as 0.003, and Adam optimizer was used for training. With a CPU of Intel Xeon(R) E5-1650 v4@3.60GHz$\times$12 and a GPU of Nvidia Titan XP, training one subject with 19 - 25 time frames or models took about seven hours. The time for testing one image was approximately 0.055 seconds. The error between the predicted point cloud and the ground truth was measured by Point Cloud-to-Point Cloud (PC-to-PC) error as:
\begin{equation}
    Error = \sum\limits_{\hat{\textbf{y}}\in\hat{\textbf{Y}}} min_{\textbf{y}\in\textbf{Y}}\vert \vert \hat{\textbf{y}} - \textbf{y}\vert \vert_2^2/ \rm numY
\end{equation}
\begin{figure}
\centering
\includegraphics[width=1.0\textwidth]{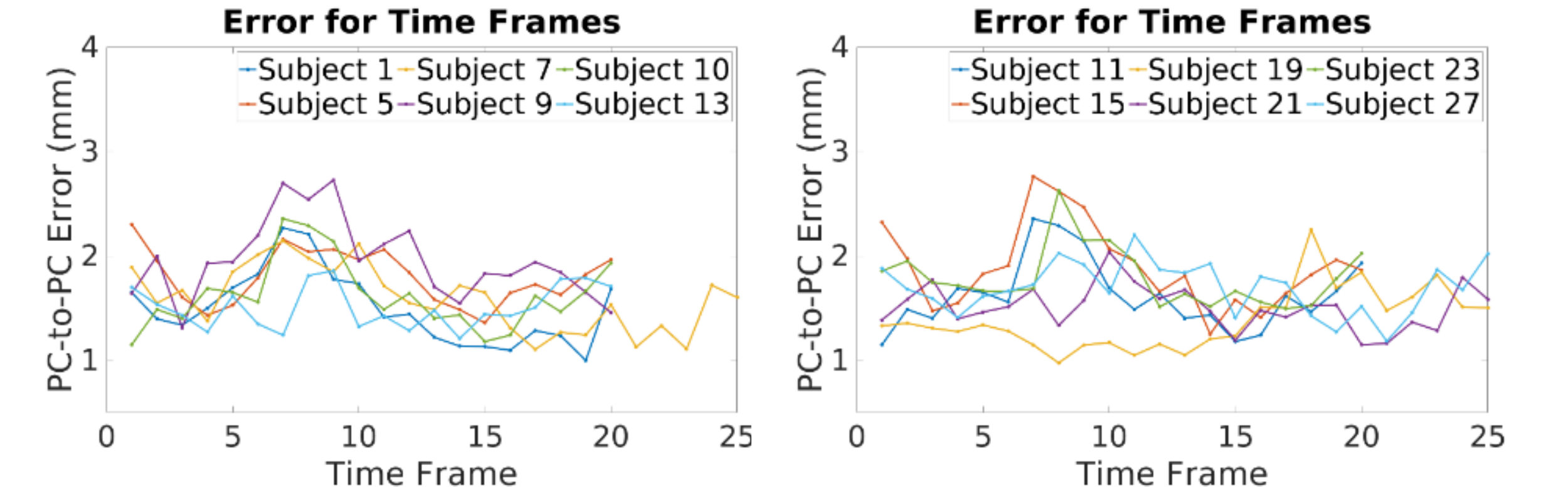}
\caption{The PC-to-PC error for each time frame for 12 subjects selected randomly from the 27 subjects.}
\label{fig:timeframe}
\end{figure}
\section{Result}
PC-to-PC errors for different time frames for 12 subjects are shown in Section \ref{sec:timeframe}. Four instantiated point clouds are shown in Section \ref{sec:example}. The result compared with the two-stage shape instantiation by using PLSR and KPLSR learning are shown in Section \ref{sec:comparison}.

\subsection{PC-to-PC Error for Each Time Frame}
\label{sec:timeframe}
The PC-to-PC errors for each time frame of 12 subjects selected randomly from the 27 subjects are shown in Figure \ref{fig:timeframe}. We can see that errors are around $1-3mm$ for all time frames. There are no excessively high peaks, which illustrates the stability of the proposed one-stage shape instantiation with PointOutNet. Slightly higher errors exist at the beginning and the end time frame (e.g. 1 and 25) and the middle time frame (e.g. 9), which is common as it was also observed in \cite{zhou2018realmed}. As these time frames are boundary time frames which are at either systole or diastole. By moving these time frames from the training data, the trained model can not see the boundaries. This happens only in this paper, as leave-one-out cross validation is used. In the real world application, this will be a minor issue, as the training data will cover all boundaries.
\begin{figure}
\centering
\includegraphics[width=1.0\textwidth]{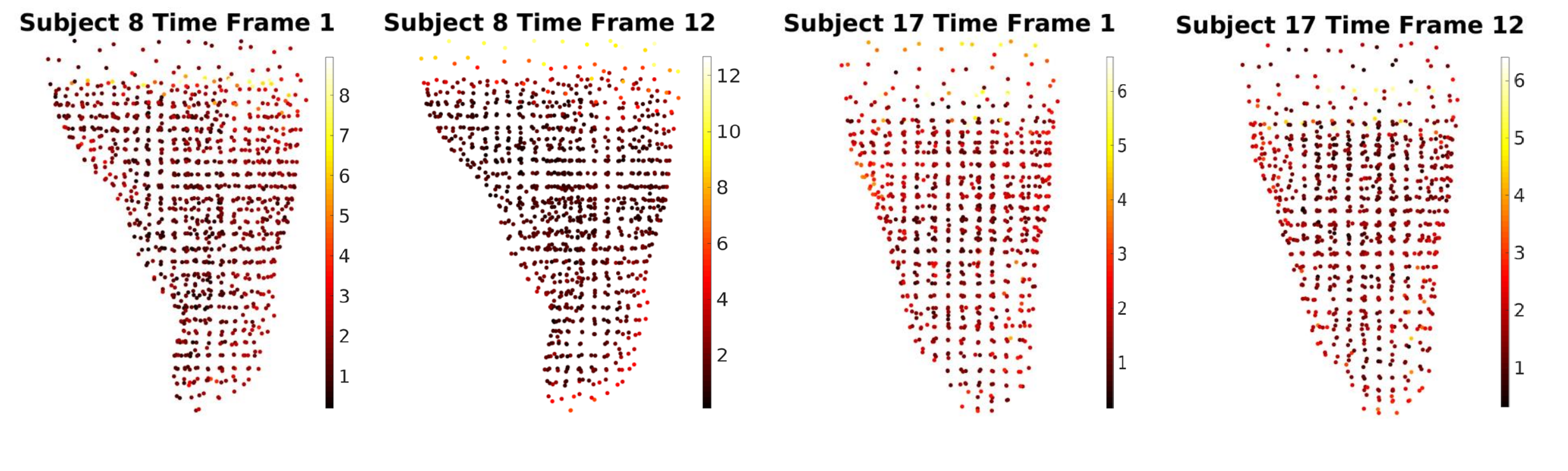}
\caption{Intuitive illustrations of the instantiation results of two randomly selected subjects at the systole and diastole time frame, color indicates the PC-to-PC error for each vertex in $mm$.}
\label{fig:example}
\end{figure}
\subsection{Instantiation Examples}
\label{sec:example}
The point clouds predicted by the PointOutNet at the systole and diastole time frames from two randomly selected subjects are shown in Figure~\ref{fig:example}. Empirically, the predictions at the systole and diastole time frames are with slightly higher errors. We can see that most vertices in the middle part have an error of $2mm$ while vertices at the top part - atrioventricular ring have higher errors. This is because the 2D MRI slices end at the atrioventricular ring, introducing plane area and sparse vertices at the top part in the 3D SSM mesh. This phenomenon increases the difficulty of vertices prediction at the top part and introduces a higher error. 
\begin{figure}
\centering
\includegraphics[width=0.8\textwidth]{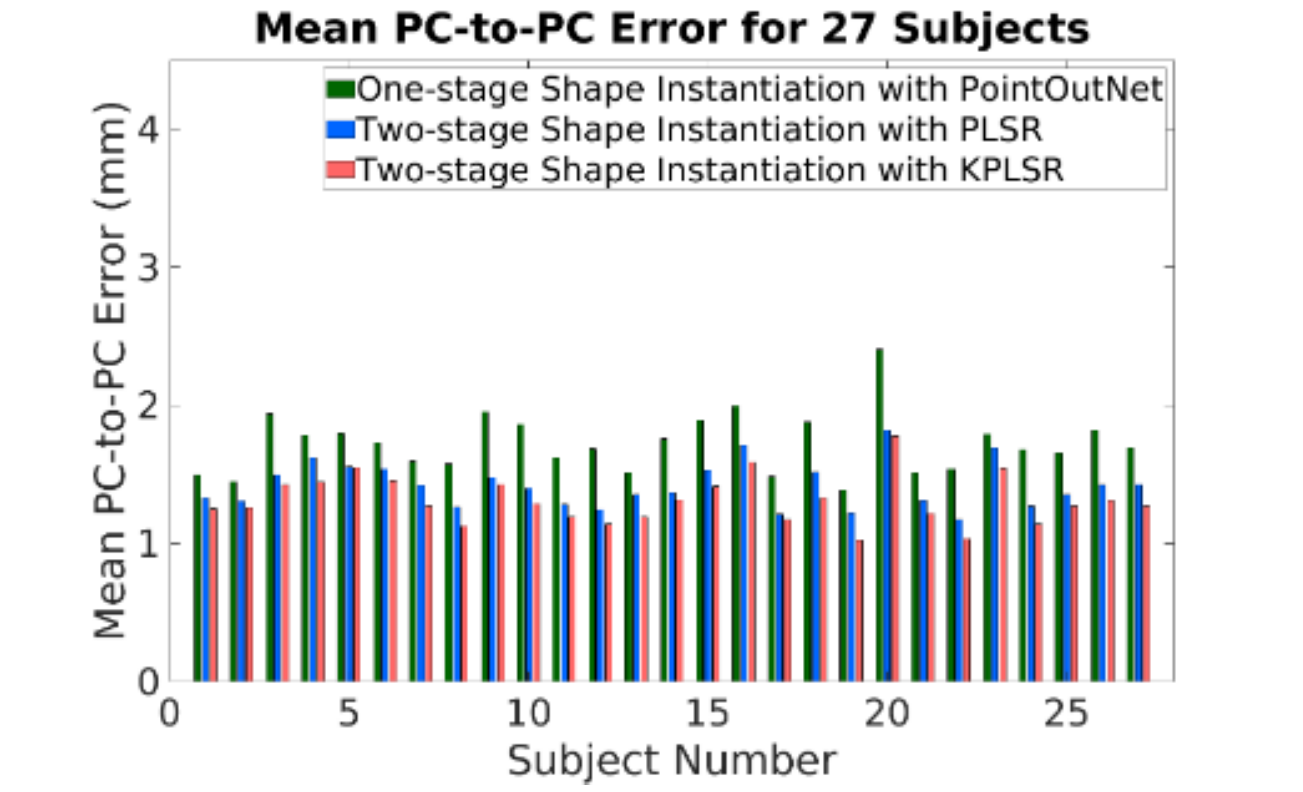}
\caption{The mean PC-to-PC error for 27 subjects with PLSR-based and KPLSR-based two-stage shape instantiation and PointOutNet-based one-stage shape instantiation.}
\label{fig:patient}
\end{figure}
\subsection{Comparisons to Other Methods}
\label{sec:comparison}
The proposed one-stage shape instantiation with PointOutNet was compared to previous two-stage shape instantiation with PLSR and KPLSR. The mean PC-to-PC error for each subject is shown in Figure~\ref{fig:patient}. We can see that comparable accuracy was achieved by the proposed method. In general, the mean PC-to-PC error of the proposed method is around $1.5mm$.

\section{Discussion}
In the one-stage shape instantiation with PointOutNet, the predicted point cloud is not with a corresponding vertex order to the ground truth. This is due to the fact that the used loss function, Chamfer distance, takes the nearest distance of a predicted vertex to the ground truth into consideration rather than the distance of a predicted vertex to a single and corresponding vertex of the ground truth. Though the Chamfer distance loss loses the vertex correspondence information, it offers larger exploration space for the network to converge. Regular L1 and L2 loss function were tested with vertex correspondences, however, the network experienced convergence difficulty and generated poor predictions. Unlike~\cite{zhou2018realmed} where the prediction is a point cloud with vertex correspondence and hence the 3D mesh can be achieved directly by applying the 3D SSM triangle connectivity, an additional 3D reconstruction algorithm is needed for a complete mesh instantiation in this paper.

Boundary effect where shape instantiation achieves worse performance at boundary time frames also exists in the one-stage shape instantiation with PointOutNet in this paper. In the real application, the training data will cover the boundary time frames and this effect will be alleviated. KPLSR-based two-stage shape instantiation has the kernel width as a sensitive hyper-parameter, which needs to be adjusted manually and carefully, whereas the proposed one-stage shape instantiation with PointOutNet is fully automatic. We think the main bottleneck for achieving higher instantiation accuracy is the loss function. We will explore this topic further in the future.

\section{Conclusion}
A one-stage shape instantiation algorithm with PointOutNet and Chamfer loss was proposed in this paper. A (2D image)-to-(3D point cloud) inference was achieved end-to-end with comparable accuracy and higher automation. To the best of the author's knowledge, this is the first work that applies deep learning into 3D shape instantiation in medical applications. It demonstrates the possibility of deep learning to achieve cross-modality tasks, which may indicate a wide application of deep learning in intra-operative and real-time navigation. For example, predicting 3D prosthesis pose or navigating medical robot's joint angle from 2D images directly.

\bibliographystyle{splncs04}
\bibliography{paper1557}
\end{document}